\documentclass{article}

\usepackage[nonatbib,final]{arxiv_nips_2017}

\usepackage[utf8]{inputenc} 
\usepackage[T1]{fontenc}    
\usepackage[colorlinks,citecolor=blue,linkcolor=blue]{hyperref}       
\usepackage{url}            
\usepackage{booktabs}       
\usepackage{amsfonts}       
\usepackage{nicefrac}       
\usepackage{microtype}      
\usepackage{verbatim}
\usepackage[svgnames]{xcolor}
\usepackage{amsmath,amssymb}
\usepackage{algorithm}
\usepackage{algorithmic}
\usepackage{graphicx}
\usepackage{subcaption}
\usepackage{tikz}
\usetikzlibrary{shapes,arrows,arrows.meta}
\tikzset{%
  >={Latex[width=2mm,length=2mm]},
            base/.style = {rectangle, rounded corners, draw=black,
                           minimum width=1.5cm, minimum height=1cm,
                           text centered, font=\sffamily},
  activityStarts/.style = {base, fill=blue!30},
    InExtractPortion/.style = {base, fill=blue!30},
       InfoAggregator/.style = {base, fill=red!30},
    ColGen/.style = {base, fill=green!10},
    Storage/.style = {base, fill=brown!30},
    LPSolve/.style = {base, fill=violet!30},
    ILPSolve/.style = {base, fill=red!30},
    Output/.style = {base, fill=gray!30},
       startstop/.style = {base, fill=red!30},
    activityRuns/.style = {base, fill=green!30},
         process/.style = {base, minimum width=2.5cm, fill=orange!15,
                           font=\ttfamily},
}
\pgfdeclarelayer{edgelayer}
\pgfdeclarelayer{nodelayer}
\pgfsetlayers{edgelayer,nodelayer,main}

\tikzstyle{vertex_default}=[circle,fill=White,draw=Black,text width=0.1cm]
\tikzstyle{tree_edge}=[thick,draw=Green]
\tikzstyle{add_edge}=[thick,draw=Red]
\tikzstyle{local_edge}=[thick,draw=Cyan]

\newcommand\ie{\emph{i.e.}}
\newcommand\eg{\emph{e.g.}}

\begin{document}
%
\title{Multi-Person Pose Estimation via Column Generation}
\author{
Shaofei Wang\\
Beijing A\&E Technologies \\
\texttt{sfwang0928@gmail.com} 
\And
Chong Zhang\\
SIMBioSys Group, DTIC, Universitat Pompeu Fabra \\
Barcelona Spain \\
\texttt{chong.zhang@upf.edu} 
\And
Miguel A. Gonzalez-Ballester\\
SIMBioSys Group, DTIC, Universitat Pompeu Fabra \& ICREA,\\
Barcelona Spain \\
\texttt{ma.gonzalez@upf.edu} 
\And
Alexander Ihler\\
University of California, Irvine \\
Irvine California \\
\texttt{ihler@ics.uci.edu} 
\And
Julian Yarkony\\
Experian Data Lab \\
San Diego CA \\
\texttt{julian.e.yarkony@gmail.com} 
}
\maketitle
\begin{abstract}
We study the problem of multi-person pose estimation in natural images.  A pose
estimate describes the spatial position and identity (head, foot, knee, etc.)
of every non-occluded body part of a person.  Pose estimation is difficult due
to issues such as deformation and variation in body configurations and occlusion
of parts, while multi-person settings add complications such as an unknown
number of people, with unknown appearance and possible interactions in their
poses and part locations.  We give a novel integer program formulation of the
multi-person pose estimation problem, in which variables correspond to
assignments of parts in the image to poses in a two-tier, hierarchical way.
This enables us to develop an efficient custom optimization procedure based on
column generation, where columns are produced by exact optimization of very
small scale integer programs.  We demonstrate improved accuracy and speed for our
method on the MPII multi-person pose estimation benchmark.
\end{abstract}

\section{Introduction}
\label{intro}

In this paper we consider the problem of multi-person pose estimation (MPPE) in
natural images. MPPE is the problem of detecting and localizing people and their
corresponding body parts.  In practice, most MPPE systems work by running part
detectors over the image, extracting a number of possible part locations, then
integrating this information using a pose model to determine both the number of
people present in the image, and the assignment of detected parts to people (the
pose).

For instance, \cite{deva3} employs a flexible mixture-of-parts model for joint
detection and estimation of human poses, where human poses are modeled by
pictorial structure \cite{felzenszwalb2005pictorial} and efficient inference is
achieved via dynamic programming and distance transform. In \cite{deva3} the
problem of finding the pose of a person is equivalent to finding the maximum a
posterior (MAP) configuration of a probabilistic graphical model where the
likelihood function  trades off two terms.  The first encourages that the part
locations of a predicted person are supported by evidence in the image as
described by local image features
\cite{dalal2005histograms,vondrick2013hoggles}.  The second encourages that the
part locations of a predicted person satisfy the angular and distance
relationships consistent with a person \cite{felzenszwalb2005pictorial}.  An
example of such a relationship is that the head of a person tends to be above
neck.  

Often, the part detectors may detect the presence of a given part several times
in close proximity, leading to a multiple detection problem; a simple way to
solve this is via non-max suppression (NMS), which removes all but the best
detections in a small region.  NMS can be done either as a pre-processing step
to suppress non-local-maximum part detections, or as a post-processing step to
suppress poses with lower scores/probabilities that overlap with poses of high
scores/probabilities. Either way, distortion or missing detection problems may
occur, particularly in multi-person images, either by removing the correct
detections, or by removing detections corresponding to separate persons.

More recent works \cite{deepcut1,deepcut2} cast the MPPE problem as an integer
linear program (ILP), in which multiple detections of a single part may be
assigned to the same person.  This allows non-max suppression to be folded into
the pose model, improving its ability to find the correct pose.  The cost
function of the ILP is generated using deep neural networks
\cite{hinton,baldi2014searching}, and the ILP is optimized using a state of the
art ILP solver, assisted by a greedy multi-stage optimization procedure.

\begin{figure}
\begin{center}
  \begin{tabular}{c c c c} 
      \includegraphics[clip,trim=.5cm 2cm 1cm 2cm,width=.21\textwidth]{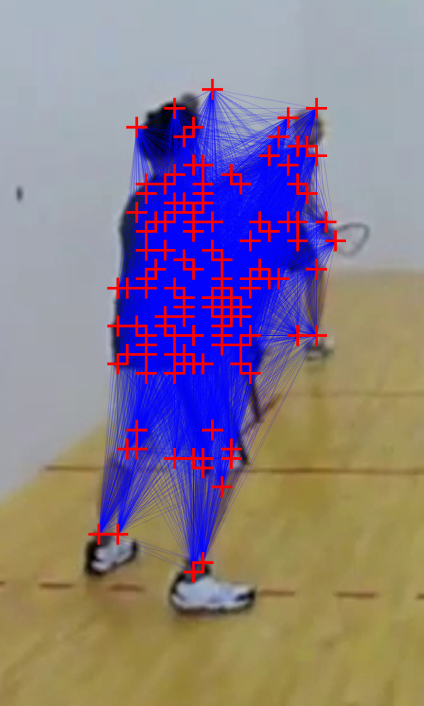} &
      \includegraphics[clip,trim=.5cm 2cm 1cm 2cm,width=.21\textwidth]{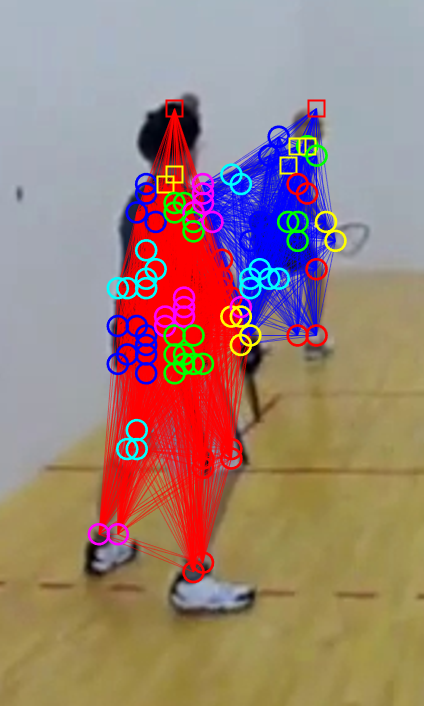} &
      \includegraphics[clip,trim=.5cm 2cm 1cm 2cm,width=.21\textwidth]{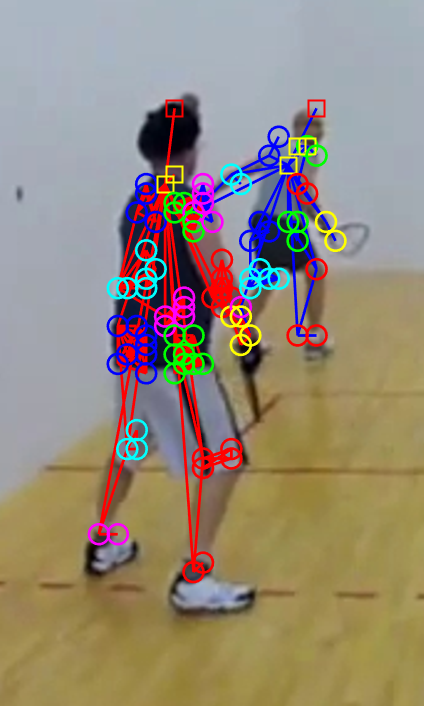} &
      \includegraphics[clip,trim=.5cm 2cm 1cm 2cm,width=.21\textwidth]{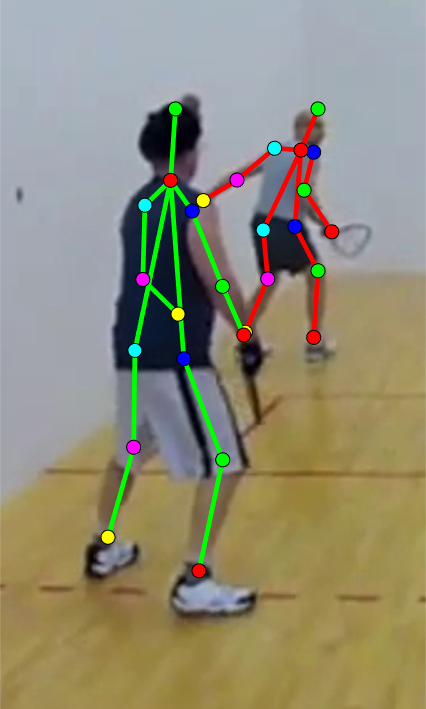}\\
      (a) raw input & (b) Deeper Cut \cite{deepcut1} & (c) our approach & (d) final
      output
  \end{tabular}
\end{center}
    \caption{Overview of our approach. (a) Raw input
    which consists of unary terms (red crosses) and pairwise terms (blue
    connections). (b) \cite{deepcut1} employs a fully-connected body model. (c)
    Our approach models the human body as an ``augmented tree'' graph. (d) We achieve
    more accurate results while being 100x faster than \cite{deepcut1}.}
  \label{fig:overview}
\end{figure}

We propose an alternative ILP formulation of MPPE,
in which we impose several additional structure assumptions on the ILP. 
In particular, we model the part assignments using a two-tier structure,
in which a local assignment tier handles non-max suppression by grouping multiple
detections, while a global pose tier handles the overall pose shape using
an augmented-tree structure for the human body.
We exploit this problem structure to design a highly efficient column generation
algorithm for optimizing the ILP \cite{cuttingstock,barnprice} tailored to this
model; for example, the global pose tier exploits the tree structured body 
model \cite{deva1,deva2,deva3}
to generate columns efficiently using dynamic programming.
Figure~\ref{fig:overview} shows an illustration contrasting \cite{deepcut1} with our
model; given many detections, \cite{deepcut1} uses a dense model to associate parts
with individuals, while our model corresponds to a two-tier structure with a tree-like body model.
In combination, this results in a novel MPPE model that is both more accurate,
and significantly faster, than the baseline method of \cite{deepcut1,deepcut2}.

We also note that a more recent approach of \cite{NL-LMP} achieves
considerable speed up over \cite{deepcut1}: it is about three orders of
magnitude faster than \cite{deepcut1} while being 10x faster than our proposed
method. Nevertheless, as will be shown later in experiments section, it is not
as accurate as our method, especially for difficult-to-localize parts such as
ankles and wrist.

Our paper is organized as follows.  In Section \ref{probForm} we outline the
assumptions of our model and its structure, then formulate it more precisely as
an ILP.  In Section \ref{violposes} we introduce our column generation approach
for computing the optimal MPPE assignment, where the column generation steps are
solved using efficient dynamic programming and small scale, exactly solvable
integer programs (IP).  In Section \ref{poseCellExp} we demonstrate that our
model and inference process provide state of the art results for MPPE on
benchmark data.  Finally, we conclude and discuss extensions in Section
\ref{concsectiondone}. Additional derivations and discussion are provided in
the supplements.

\section{Multi-Person Pose Estimation Model}
\label{probForm}
In this section, we describe our two-tiered structure for reasoning about pose
estimation.  The input to our model is a set of body part detections; in
practice, we use the body part detector of \cite{deepcut1}, which employs a deep
convolutional neural network \cite{Deepseg,krizhevsky2012imagenet}.  Each
detection is associated with exactly one body part.  Our model uses fourteen
parts, consisting of the head and neck, along with right and left variants of
the ankle, knee, hip, wrist, elbow, and shoulder.  We use the term
\emph{complete pose} to describe a person in an image, as represented by the
detections associated with their body parts.

\subsection{Assignment of Parts and Validity}
\label{keyPointCombRules}
We partition the body parts into two types: \emph{major} parts, of which at
least one is required to be present (not occluded) in any complete pose, and
\emph{minor} parts, any of which may be occluded.  In practice, we take the neck
to be the only major part, thus requiring that each complete pose be associated
with at least one neck detection.  

We reason about the assignment of parts to a complete pose in two tiers: a
\emph{local assignment}, which corresponds to a grouping of detections for a
single part that are all associated with a single complete pose; and a
\emph{global pose}, which corresponds to at most one detection of each part.  In
practice, the score of a local assignment evaluates the coherence of the
detections for that part (for example, two visually similar detections of a part
in close proximity are more likely to correspond to the same person), while the score
of the global pose captures the coherence of these part locations according to a
(nearly) tree structured model of the human body (for example, the head is
typically located above the neck).  In any local assignment, we require that
exactly one detection be assigned to some global pose, so that the global pose
reasons about the overall position and visibility of the person, and the local
assignment captures any additional detections associated with each visible part.
A \emph{complete pose} corresponds to a single person in the image, and consists
of a single global pose and the local assignments (additional detections)
associated with each of its visible parts.

Finally, we categorize detections as either \emph{global}, \emph{local}, or
\emph{false positive}.  Global detections are those associated with some global
pose; local detections are the non-global detections in a local assignment; and
false positives are detections not contained in any global pose or local
assignment.

These definitions result in the following requirements for a set of complete
poses, which describe a group of people in the image:
\begin{enumerate}
  \item A detection can only be global, local, or neither.
  \item No two global poses can share a common detection.
  \item No two local assignments can share a common detection.
  \item The global detection of a local assignment must also be included in a
        global pose.
\end{enumerate}
We refer to these conditions as the \emph{validity conditions} and a selection of
global poses and local assignments that meet them is referred to as \emph{valid}.  
\subsection{Integer Linear Program Formulation}
\label{MathForm}

\begin{table*}[t] \centering
\resizebox{\textwidth}{!}{
\begin{tabular}{|| l l l l ||} 
 \hline
 Term & Form &Index & Meaning  \\ [0.5ex] 
 \hline\hline
 $\mathcal{D}$& set & $d$ & set of detections  \\ 
 $\mathcal{R}$& set  & $r$ & set of parts  \\
 $\mathcal{R}'$& set & $r$ & set of major parts; $\mathcal{R}' \subseteq \mathcal{R}$  \\
 $R $ & $\{0,1\} ^{|\mathcal{D}|\times |\mathcal{R}|}$ &  $R_{dr}$,$R_d$ & $R_{dr}=1$ indicates that detection $d$ is associated with part $r$.    \\
 $R_d $& $R_d \in \mathcal{R}$  & none & short hand for $\mbox{arg}\max_{r}R_{dr}$ \\ 
 $\mathcal{G}$ & set & $q$ & set of all global poses\\ 
 $\mathcal{L}$& set & $l$ & set of all local assignments \\ 
  $\hat{\mathcal{G}}$ & set & $q$ & set of  global poses generated during column generation\\ 
 $\hat{\mathcal{L}}$& set & $l$ & set of local assignments generated during column generation \\ 
$\theta $ & $ \mathbb{R}^{|\mathcal{D}|}$ & $d$ & $\theta_d$ is the cost of including $d$ in a complete pose\\
$\phi $ & $ \mathbb{R}^{|\mathcal{D}|\times |\mathcal{D}|}$ & $d_1,d_2$ & $\phi_{d_1d_2}$ is the cost of including $d_1,d_2$ in the same local assignment or global pose\\
 $G $ &  $\{0,1\}^{|\mathcal{D}|\times |\mathcal{G}|}$ & $d,q$ & $G_{dq}=1$ indicates that $d$ is a global detection in global pose $q$\\
 $L $ &  $\{0,1\}^{|\mathcal{D}|\times |\mathcal{L}|}$ & $d,l$ & $L_{dl}=1$ indicates that $d$ is a local detection in local assignment $l$\\
 $M $ &  $\{0,1\}^{|\mathcal{D}|\times |\mathcal{L}|}$ & $d,l$ & $M_{dl}=1$ indicates that $d$ is a global detection in local assignment $l$\\
 $\Gamma$ & $\mathbb{R}^{|\mathcal{G}|}$ & $q$ & $\Gamma_q$ is the cost of global pose $q$\\
 $\Psi$ & $\mathbb{R}^{|\mathcal{L}|}$ & $l$ & $\Psi_l$ is the cost of local assignment $l$\\
 $\gamma$ & $\{ 0,1\}^{|\mathcal{G}|}$ & $q$ & $\gamma_q=1$ indicates that global pose $q$ is selected.\\
 $\psi$ & $\{ 0,1\}^{|\mathcal{L}|}$ & $l$ & $\psi_l=1$ indicates that local assignment $l$ is selected.
 \\
 [1ex]
 \hline
\end{tabular}
}
\caption{Summary of Notation}
\label{myNotationTable}
\end{table*}

We now formally define the MPPE task as an integer linear program (ILP).  We
first describe the variables associated with detections and parts, global poses,
and local assignments; give the validity constraints on these variables as
linear inequalities; and finally define the cost of a pose and the overall
optimization problem, and discuss its linear program (LP) relaxation.  We
summarize our notation in Table \ref{myNotationTable}.

\paragraph{Detections and Parts.}
We denote the set of detections in the image as $\mathcal{D}$, and index these detections by $d$.
Similarly, we use $\mathcal{R}$ to denote the set of parts, indexed by $r$, and denote the
set of major parts by $\mathcal{R}' \subseteq \mathcal{R}$.
We describe the mapping of detections to parts using a matrix $R \in \{ 0,1
\}^{|\mathcal{D}|  \times |\mathcal{R}|}$, indexed by $d,r$.  
Specifically, $R_{dr}=1$ 
indicates that detection $d$ is associated with part $r$.  As a useful shorthand,
we define $R_{d}$ to be the part associated with detection $d$.  
\paragraph{Global Poses.}
%
Given the set of detections $\mathcal{D}$, we define the set of all possible global poses
over $\mathcal{D}$ as $\mathcal{G}$. Members of $\mathcal{G}$ have at least one
global detection corresponding to a major part and no more than one detection
corresponding to any given part.  We describe mappings of detections to global
poses using a matrix $G \in \{ 0,1\}^{|\mathcal{D}|\times |\mathcal{G}|}$, and set 
$G_{dq}=1$ if and only if detection $d$ is associated with global pose $q$. 

Note that the set of all possible poses $\mathcal{G}$ is impractically large (it
contains all valid assignments of detections to a global pose).  Thus in
practice, we never construct $G$ explicitly; instead, we maintain an active set
of poses, $\hat{\mathcal{G}}$, restricting $G$ to this set.

\paragraph{Local Assignments.}
%
Next we denote
the set of all possible local assignments
over the detections $\mathcal{D}$ by $\mathcal{L}$, and index these possible
local assignments by $l$.  
Since we require that, for any local assignment $l\in \mathcal{L}$, exactly
one of the detections in $l$ is global, we describe $\mathcal{L}$ using
two matrices $L, M \in \{0,1\}^{|\mathcal{D}| \times |\mathcal{L}|}$,
where $L_{dl}=1$ if and only if detection $d$ is associated with $l$ as a local (non-global)
detection, and $M_{dl}=1$ if and only if detection $d$ is associated with
$l$ as a global detection. 

The set $\mathcal{L}$ is too large to be considered explicitly during
optimization.  We maintain a subset $\hat{\mathcal{L}} \subseteq \mathcal{L}$
during optimization, and explictly represent $L$ and $M$ restricted to
$\hat{\mathcal{L}}$.

\paragraph{Validity Constraints.}
\label{validDef}
We index a set of global poses and local assignments using indicator vectors,
so that $\gamma \in \{0,1\}^{|\mathcal{G}|}$ with 
$\gamma_q=1$ to indicate that global pose $q \in \mathcal{G}$ is selected, and 
$\gamma_q=0$ otherwise.  Similarly, we let $\psi\in \{0,1\}^{|\mathcal{L}|}$
with $\psi_l=1$ to indicate that local
assignment $l\in \mathcal{L}$ is selected, with $\psi_l=0$ otherwise.

A solution $\gamma,\psi$ is a valid solution if and only if it satisfies the rules defined previously, which is written formally as the following set of linear inequalities:
\begin{align*}
G\gamma+L\psi &\leq 1 & &:\ \mbox{A detection can only be global, local, or neither;}\\
& & &\qquad \mbox{no two global poses can share the same detection.}\\
L\psi+M\psi &\leq 1   & &:\ \mbox{No two local assignments can share the same detection.} \\
-G\gamma+M\psi&\leq 0 & &:\ \mbox{The global detection of a local assignment is included in some global pose.}
\end{align*}

\paragraph{Cost Function.}
\label{costPose}
We now describe the cost function for MPPE.  
Our total cost is expressed in terms of unary costs $\theta \in \mathbb{R}^{|\mathcal{D}|}$,
where $\theta_d$ is the cost of assigning detection $d$ to a pose,
and pairwise costs $\phi \in \mathbb{R}^{|\mathcal{D}|\times |\mathcal{D}|}$, 
where $\phi_{d_1 d_2}$ is the cost of assigning
detections $d_1$ and $d_2$ to a common global pose or local assignment. 
We use $\omega$ to
denote the cost of instancing a pose, which serves to regularize the
number of people in an image.  

The cost of a complete pose is thus the sum of the costs of the following.  
\begin{itemize}
\item $\phi$ terms associated with pairs of detections in its global pose 
\item $\phi$ terms associated with pairs of detections within each of its local
    assignments
\item $\theta$ terms associated with detections in either its global or local assignments
\item $\omega$ term associated with instancing a pose.  
\end{itemize}
For convenience, we separate these costs into $\Gamma_q$ as the cost associated with the global pose
$q$, and $\Psi_{l}$ as the cost of local assignment $l$, respectively:
\begin{align*}
\Gamma_q&=\omega+\sum_{d\in \mathcal{D}}\theta_dG_{dq}+\sum_{\substack{d_1\in \mathcal{D}\\d_2\in \mathcal{D}}}\phi_{d_1d_2}G_{d_1q}G_{d_2q}\\
\Psi_{l}&=\sum_{d\in \mathcal{D}}\theta_d L_{dl} 
 + \sum_{\substack{d_1\in \mathcal{D}\\d_2\in \mathcal{D}}}\phi_{d_1d_2}(L_{d_1 l}+M_{d_1 l})(M_{d_2 l}+L_{d_2 l})
\end{align*}
%
\paragraph{Integer Linear Program.}
\label{optDef}
We now cast the problem of finding the lowest cost set of poses as an
integer linear program subject to our validity constraints:
\begin{align}
\label{primdualilp}
\min_{\substack{\gamma \in \{0,1\}^{|\mathcal{G}|} \\ \psi \in \{0,1\}^{|\mathcal{L}|}}} & \Gamma^\top\gamma+\Psi^\top\psi 
&& \text{s.t.} 
&& G\gamma+L\psi \leq 1\quad 
&&  L\psi+M\psi\leq 1
&& -G\gamma+M\psi\leq 0
\end{align}
By relaxing the integrality constraints on $\gamma,\psi$, we obtain a linear program relaxation
of the ILP, and can convert Eq.~\eqref{primdualilp} to its dual form using Lagrange multiplier sets
$\lambda^1,\lambda^2,\lambda^3 \in \mathbb{R}_{0+}^{|\mathcal{D}|}$:  

\begin{align}
\label{dualFormPose}
\min_{\substack{\gamma\geq 0 \\ \psi \geq 0 \\ G\gamma+L\psi \leq 1 \\ L\psi+M\psi\leq 1 \\-G\gamma+M\psi\leq 0}}\Gamma^\top\gamma+\Psi^\top \psi 
 =\max_{\substack{\lambda^1\geq 0\\ \lambda^2\geq 0 \\ \lambda^3\geq 0 \\  \Gamma+G^\top(\lambda^1-\lambda^3) \geq 0 \\ \Psi+L^\top\lambda^1+(M^\top+L^\top)\lambda^2+M^\top\lambda^3 \geq 0}} -1^\top\lambda^1-1^\top\lambda^2
\end{align}

\section{Column Generation Solution}
\label{violposes}

In this section we consider optimization of the LP relaxation in Eq.~\eqref{dualFormPose}.
As discussed, the primary difficulty is the intractable sizes of the sets $\mathcal{G},\mathcal{L}$.
Instead, we consider subsets $\hat{\mathcal{G}} \subseteq \mathcal{G}$ and 
$\hat{\mathcal{L}} \subseteq \mathcal{L}$ that are constructed
strategically during optimization so as to be small, while still solving the LP
in Eq.~\eqref{dualFormPose} exactly.
This type of column generation approach is common in the operations research literature,
in which the task of generating the columns is often called \emph{pricing} \cite{barnprice}.

\begin{figure}[t]
\centering
\begin{subfigure}{.75\textwidth}
\resizebox{\textwidth}{!}{
\begin{tikzpicture}[node distance=2cm,
    every node/.style={fill=white, font=\sffamily}, align=center]
%
 \node (rawImage)         [InExtractPortion]              {Image}; \\
 \node (DeepNet)          [InExtractPortion,left of =rawImage,xshift=-.001cm]              {Deep Net};\\
 \node (CRFGen)           [InExtractPortion,left of =DeepNet,xshift=-1cm]              {Cost Generator};
 \node (CostAgg)          [ColGen,left of =CRFGen,yshift=-.1cm,xshift=-1cm]              {$\theta,\phi,\lambda$};
 \node (Col1)             [ColGen,below of =CostAgg,xshift=-2cm]              {Opt Local};
 \node (Col2)             [ColGen,below of =CostAgg,xshift=2cm]              {Opt Global};
 \node (hatP)             [ColGen,below of =Col2,xshift=-2cm]              {$\hat{\mathcal{G}},\hat{\mathcal{L}}$};
 \node (LPSolve)          [Storage,left of =hatP,xshift=-3cm]              {Dual LP};
 \node (ILPSolve)         [ILPSolve,right of =LPSolve,xshift=7cm]              {Primal ILP};
 \node (OutputModel)      [ILPSolve,right of =ILPSolve,xshift=1.5cm]              {Output\\ to User};
 \draw[->]                (rawImage) --  (DeepNet);
 \draw[->]                (DeepNet) -- node {$\mathcal{D}$}(CRFGen);
 \draw[->]                (CRFGen) --  (CostAgg);
 \draw[->]                (CostAgg) -- (Col1.north);
 \draw[->]                (CostAgg) -- (Col2.north);
 %
   \draw[->]              (Col1.south) -- node {$\dot{\mathcal{L}}$}(hatP);
 \draw[->]                (Col2.south) -- node {$\dot{\mathcal{G}}$}(hatP);

 \draw[->]                (hatP) -- node {$\hat{\mathcal{G}},\hat{\mathcal{L}}$}(LPSolve);
 \draw[->]                (hatP) -- node {$\hat{\mathcal{G}},\hat{\mathcal{L}}$}(ILPSolve);
 \draw[->]                (ILPSolve) -- node {$\gamma,\psi$}(OutputModel);
 \draw[->]                (LPSolve.north) -- ++(0,0) --  ++(0,3.5) -- ++(0,0)--node {$\lambda$}(CostAgg.west);
\end{tikzpicture}
}
\end{subfigure}
\caption{Diagram of our system: blue blocks represent steps for generating unary
  and pairwise costs, which are identical to that of \cite{deepcut1}.  Cost
  generator is the procedure for mapping the output scores of the deep neural
  network to unary cost terms and computing pairwise costs based on geometric
  features.  Green blocks represent steps for generating columns.  \textbf{Opt
  Local} and \textbf{Opt Global} correspond to the pricing problems in line 5-10
  and line 11-16 of Alg \ref{dualsolvesimpless}, respectively.  The brown block
  represents a dual LP solver while red blocks show steps for producing the final
  integer solutions at termination.}
\label{ColGenDescFig}
\end{figure}
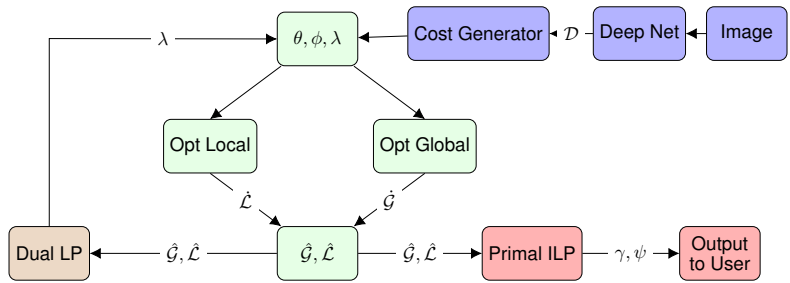

We solve the dual form LP in Eq.~\eqref{dualFormPose} iteratively with two
steps. We first solve the dual LP over constraint sets
$\hat{\mathcal{G}}$ and $\hat{\mathcal{L}}$, which are initialized to be empty.
Then, we identify violated constraints in the dual using 
combinatorial optimization and add these to sets $\hat{\mathcal{G}}$ and
$\hat{\mathcal{L}}$.  One local assignment is identified corresponding to each possible selection of a global detection, and one global pose is identified for each selection of a
detection corresponding to a major part.  We repeat these two steps until no more
violated constraints exist.  We then solve the integer linear program
over sets $\hat{\mathcal{G}}$ and $\hat{\mathcal{L}}$.  We diagram this procedure in Figure
\ref{ColGenDescFig} and show the corresponding algorithm in Alg ~\ref{dualsolvesimpless}.

\begin{algorithm}[t]
\caption{Dual Optimization}
\begin{algorithmic}[1]
\STATE $\hat{\mathcal{G}} \leftarrow \{ \}$
\STATE $\hat{\mathcal{L}} \leftarrow \{ \}$
\REPEAT
\STATE $\lambda \leftarrow$ Maximize dual in Eq.~\eqref{dualFormPose} over column sets $\hat{\mathcal{G}},\hat{\mathcal{L}}$
\FOR {$d_* \in \mathcal{D}$}
\STATE $l_* \leftarrow \mbox{arg} \min_{\substack{l\in \mathcal{L} \\ M_{d_*l}=1}}(\lambda^3_{d_*}+\lambda^2_{d_*})M_{d_*l}+\sum_{d \in \mathcal{D}}(\lambda^1_d+\lambda^2_d)L_{dl}+\Psi_l$
\IF{$(\lambda^3_{d_*}+\lambda^2_{d_*})M_{d_*l_*}+\sum_{d \in \mathcal{D}}(\lambda^1_d+\lambda^2_d)L_{dl_*}+\Psi_{l_*}<0$}
\STATE $\dot{\mathcal{L}} \leftarrow [\dot{\mathcal{L}} \cup l_*]$
\ENDIF
\ENDFOR
\FOR {$d_{*} \in \mathcal{D}$  s.t. $ R_{d_*}$  $ \in \mathcal{R}'$}
\STATE $q_* \leftarrow \mbox{arg} \min_{\substack{q \in \mathcal{G} \\ G_{d_*q}=1}}\Gamma_q+\sum_{d \in \mathcal{D}} G_{dq}(\lambda^1_d-\lambda^3_d)$
\IF{$\Gamma_{q_*}+\sum_{d \in \mathcal{D}} G_{dq_*}(\lambda^1_d-\lambda^3_d)<0$}
\STATE $\dot{\mathcal{G}} \leftarrow [\dot{\mathcal{G}} \cup q_*]$
\ENDIF
\ENDFOR
\STATE  $\hat{\mathcal{L}}\leftarrow [\hat{\mathcal{L}},\dot{\mathcal{L}}]$
\STATE  $\hat{\mathcal{G}}\leftarrow [\hat{\mathcal{G}},\dot{\mathcal{G}}]$
 \UNTIL{ $|\dot{\mathcal{G}}|+|\dot{\mathcal{L}}|  =0 $}
\end{algorithmic}
\label{dualsolvesimpless}
\end{algorithm}

\subsection{Identifying Violated Local Assignments}
\label{mostviolloal}
For each detection $d_*\in \mathcal{D}$, we compute the most violated constraint
corresponding to a local assignment in which $d_*$ is the global detection.  We write
this as an IP using the indicator vector $x\in \{0,1\}^{|\mathcal{D}|}$, and
define a new column $l$ for inclusion in matrices $L$ and $M$,
assigning $M_{d^*l}=1$ and $L_{dl}=x^*_d$ for all $d\in \mathcal{D}$, where $x^*$
is the solution to

\begin{align}
\label{mostViolGlobalEqu}
& \min_{\substack{l\in \mathcal{L} \\ M_{d_*l}=1}}(\lambda^2_{d_*}+\lambda^3_{d_*})M_{d_*l}+\sum_{d \in \mathcal{D}}(\lambda^1_d+\lambda^2_d)L_{dl}+\Psi_l \nonumber \\
= & \min_{\substack{x \in \{ 0,1\}^{|\mathcal{D}|} \\x_{d_*}=1 \\ x_d = 0 \; \; \forall R_d \neq R_{d_*}}}
(\lambda^2_{d_*}+\lambda^3_{d_*}) +\sum_{d \in \mathcal{D} \setminus d_*} (\theta_d+\lambda^1_{d}+\lambda^2_d)x_d + \sum_{d_1,d_2 \in \mathcal{D}} x_{d_1} x_{d_2} \phi_{d_1d_2}
\end{align}
In practice, we solve this IP by explicit enumeration over the possible local assignments.
Since the number of detections associated with any given part (and thus eligible to participate
in the local assignment of $d^*$) is small -- no larger than 15 and usually less than 10 --
exhaustive search is feasible.  One can convert this problem to an equivalent ILP problem and
use an off-the-shelf ILP solver that employs branch-and-cut to solve it.

\subsection{Identifying Violated Global Poses}
\label{mostviolglobal}
For each detection $d_*$ such that $R_{d_*} \in \mathcal{R}'$ (\ie, $d_*$ corresponds
to a major part), we compute the most violated constraint corresponding to a
global pose that includes detection $d_*$.  Again, we write this as an IP using an indicator vector
$x\in \{0,1\}^{|\mathcal{D}|}$, and define a new column $q$ to be included in $G$,
defined by 
$G_{dq} = x^*_d$ for all $d \in \mathcal{D}$, where $x^*$ is the solution to:
\begin{equation}
\min_{\substack{q \in \mathcal{G} \\ G_{d_*q}=1}} \Gamma_q+\sum_{d \in \mathcal{D}} G_{dq}(\lambda^1_d-\lambda^3_d)
 \quad = \quad 
\omega + \!\!\!\!\!\! \min_{ \substack{x \in \{ 0,1\}^{|\mathcal{D}|} \\ x_{d_*}=1\\  \sum_{d \in \mathcal{D}}R_{dr}x_d \leq 1 \; \; \forall r \in \mathcal{R}}} \!\!\!\!\!\!\! \sum_{ d\in \mathcal{D}} (\theta_{d}+\lambda^1_d-\lambda^3_d)x_{d}
 +\sum_{d_1d_2 \in \mathcal{D}} \phi_{d_1d_2}x_{d_1}x_{d_2}
\end{equation}

By enforcing some structure in the pairwise costs $\phi$, we can ensure that this optimization
problem is tractable.  A common model in computer vision is to represent the location
of parts in the body using a tree-structured model, for example in the deformable part model
of \cite{deva1,deva2,deva3}; this forces the $\phi$ terms to be zero between non-adjacent
parts on the tree.

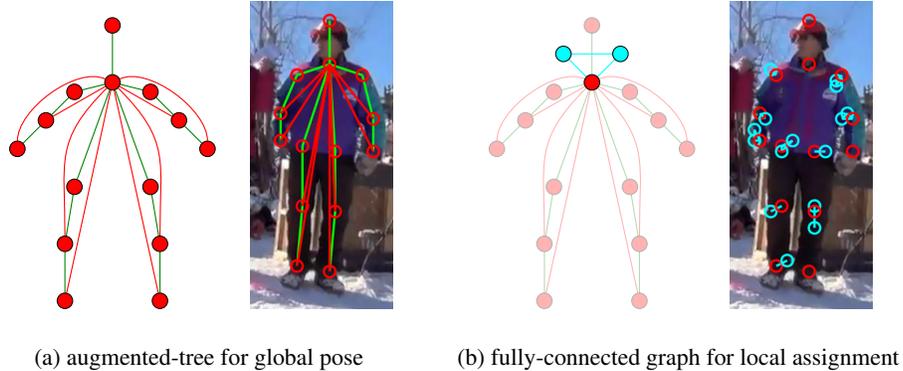
\begin{figure}[t]
\begin{center}
\begin{subfigure}[b]{0.45\textwidth}
\begin{center}
\begin{tabular}{ c c } 
\resizebox{0.45\textwidth}{!}{
\begin{tikzpicture}
	\begin{pgfonlayer}{nodelayer}
		\node [style=vertex_default, fill=Red] (0) at (-3.75, 5) {};      
		\node [style=vertex_default, fill=Red] (1) at (-3.75, 3.5) {};    
		\node [style=vertex_default, fill=Red] (2) at (-4.75, 3.25) {};   
		\node [style=vertex_default, fill=Red] (3) at (-2.75, 3.25) {};   
		\node [style=vertex_default, fill=Red] (4) at (-5.5, 2.5) {};     
		\node [style=vertex_default, fill=Red] (5) at (-2, 2.5) {};       
		\node [style=vertex_default, fill=Red] (6) at (-6.25, 1.75) {};   
		\node [style=vertex_default, fill=Red] (7) at (-1.25, 1.75) {};   
		\node [style=vertex_default, fill=Red] (8) at (-4.75, 0.75) {};   
		\node [style=vertex_default, fill=Red] (9) at (-2.75, 0.75) {};   
		\node [style=vertex_default, fill=Red] (10) at (-5, -0.75) {};    
		\node [style=vertex_default, fill=Red] (11) at (-2.5, -0.75) {};  
		\node [style=vertex_default, fill=Red] (12) at (-5, -2.25) {};    
		\node [style=vertex_default, fill=Red] (13) at (-2.5, -2.25) {};  
	\end{pgfonlayer}
	\begin{pgfonlayer}{edgelayer}
		\draw [style=tree_edge] (0) to (1);
		\draw [style=tree_edge] (1) to (2);
		\draw [style=tree_edge] (1) to (3);
		\draw [style=tree_edge] (2) to (4);
		\draw [style=tree_edge] (3) to (5);
		\draw [style=tree_edge] (4) to (6);
		\draw [style=tree_edge] (5) to (7);
		\draw [style=tree_edge] (1) to (8);
		\draw [style=tree_edge] (1) to (9);
		\draw [style=tree_edge] (8) to (10);
		\draw [style=tree_edge] (9) to (11);
		\draw [style=tree_edge] (10) to (12);
		\draw [style=tree_edge] (11) to (13);

		\draw [style=add_edge] (1) to (4);
		\draw [style=add_edge] (1) to (5);
		\draw [style=add_edge, in=90, out=155, looseness=.75] (1) to (6);
		\draw [style=add_edge, in=90, out=25, looseness=.75] (1) to (7);
		\draw [style=add_edge, in=90, out=225, looseness=1.25] (1) to (10);
		\draw [style=add_edge, in=90, out=315, looseness=1.25] (1) to (11);
		\draw [style=add_edge] (1) to (12);
		\draw [style=add_edge] (1) to (13);
	\end{pgfonlayer}
\end{tikzpicture}
} &
\includegraphics[clip,trim=10cm 1cm 2cm 2cm,width=.3\textwidth]{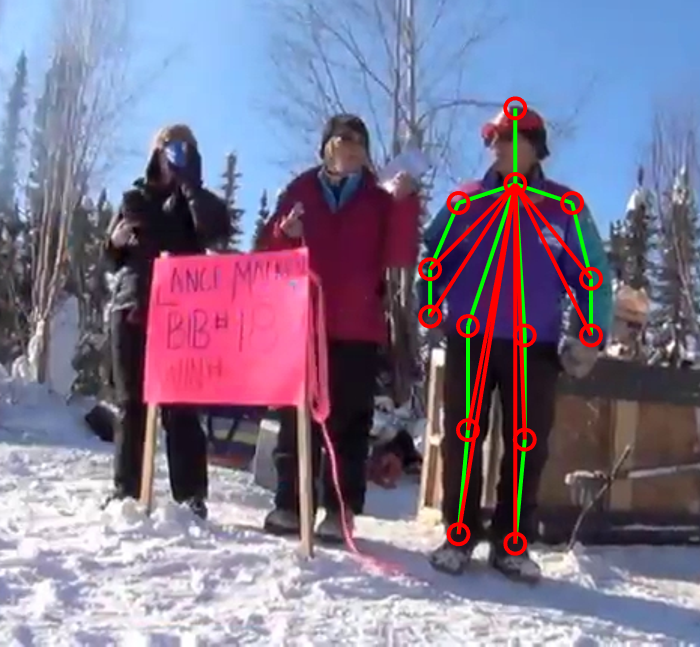}
\end{tabular}
\end{center}
\caption{augmented-tree for global pose}
\end{subfigure}
\begin{subfigure}[b]{0.45\textwidth}
\begin{center}
\begin{tabular}{ c c } 
\resizebox{0.45\textwidth}{!}{
\begin{tikzpicture}
	\begin{pgfonlayer}{nodelayer}
		\node [style=vertex_default, draw opacity=.3, fill=Red, fill opacity=.3] (0) at (-3.75, 5) {};      
		\node [style=vertex_default, fill=Red] (1) at (-3.75, 3.5) {};    
		\node [style=vertex_default, draw opacity=.3, fill=Red, fill opacity=.3] (2) at (-4.75, 3.25) {};   
		\node [style=vertex_default, draw opacity=.3, fill=Red, fill opacity=.3] (3) at (-2.75, 3.25) {};   
		\node [style=vertex_default, draw opacity=.3, fill=Red, fill opacity=.3] (4) at (-5.5, 2.5) {};     
		\node [style=vertex_default, draw opacity=.3, fill=Red, fill opacity=.3] (5) at (-2, 2.5) {};       
		\node [style=vertex_default, draw opacity=.3, fill=Red, fill opacity=.3] (6) at (-6.25, 1.75) {};   
		\node [style=vertex_default, draw opacity=.3, fill=Red, fill opacity=.3] (7) at (-1.25, 1.75) {};   
		\node [style=vertex_default, draw opacity=.3, fill=Red, fill opacity=.3] (8) at (-4.75, 0.75) {};   
		\node [style=vertex_default, draw opacity=.3, fill=Red, fill opacity=.3] (9) at (-2.75, 0.75) {};   
		\node [style=vertex_default, draw opacity=.3, fill=Red, fill opacity=.3] (10) at (-5, -0.75) {};    
		\node [style=vertex_default, draw opacity=.3, fill=Red, fill opacity=.3] (11) at (-2.5, -0.75) {};  
		\node [style=vertex_default, draw opacity=.3, fill=Red, fill opacity=.3] (12) at (-5, -2.25) {};    
		\node [style=vertex_default, draw opacity=.3, fill=Red, fill opacity=.3] (13) at (-2.5, -2.25) {};  
		\node [style=vertex_default, fill=Cyan] (14) at (-4.5, 4.25) {};    
		\node [style=vertex_default, fill=Cyan] (15) at (-3, 4.25) {};  
	\end{pgfonlayer}
	\begin{pgfonlayer}{edgelayer}
		\draw [style=tree_edge, draw opacity=.3] (0) to (1);
		\draw [style=tree_edge, draw opacity=.3] (1) to (2);
		\draw [style=tree_edge, draw opacity=.3] (1) to (3);
		\draw [style=tree_edge, draw opacity=.3] (2) to (4);
		\draw [style=tree_edge, draw opacity=.3] (3) to (5);
		\draw [style=tree_edge, draw opacity=.3] (4) to (6);
		\draw [style=tree_edge, draw opacity=.3] (5) to (7);
		\draw [style=tree_edge, draw opacity=.3] (1) to (8);
		\draw [style=tree_edge, draw opacity=.3] (1) to (9);
		\draw [style=tree_edge, draw opacity=.3] (8) to (10);
		\draw [style=tree_edge, draw opacity=.3] (9) to (11);
		\draw [style=tree_edge, draw opacity=.3] (10) to (12);
		\draw [style=tree_edge, draw opacity=.3] (11) to (13);

		\draw [style=add_edge, draw opacity=.3] (1) to (4);
		\draw [style=add_edge, draw opacity=.3] (1) to (5);
		\draw [style=add_edge, in=90, out=155, looseness=.75, draw opacity=.3] (1) to (6);
		\draw [style=add_edge, in=90, out=25, looseness=.75, draw opacity=.3] (1) to (7);
		\draw [style=add_edge, in=90, out=225, looseness=1.25, draw opacity=.3] (1) to (10);
		\draw [style=add_edge, in=90, out=315, looseness=1.25, draw opacity=.3] (1) to (11);
		\draw [style=add_edge, draw opacity=.3] (1) to (12);
		\draw [style=add_edge, draw opacity=.3] (1) to (13);

		\draw [style=local_edge] (1) to (14);
		\draw [style=local_edge] (1) to (15);
		\draw [style=local_edge] (14) to (15);
	\end{pgfonlayer}
\end{tikzpicture}
} &
\includegraphics[clip,trim=10cm 1cm 2cm 2cm,width=.3\textwidth]{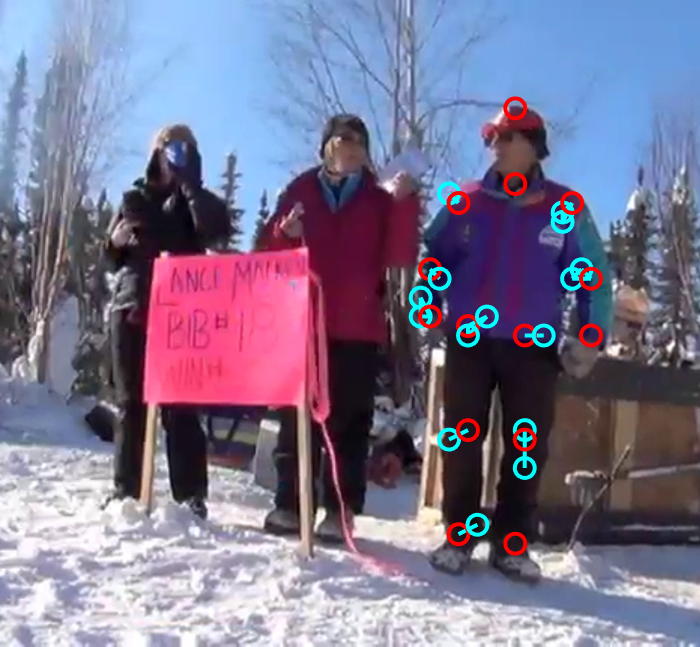}
\end{tabular}
\end{center}
\caption{fully-connected graph for local assignment}
\end{subfigure}
\end{center}
\caption{Graphical representation of our pose model. (a) A global pose is
modeled by an augmented-tree, in which each red node represents a global
detection, green edges are connections of traditional pictorial structure, while
red edges are augmented connections from neck to all non-adjacent parts of neck.
(b) Each local assignment is modeled by a fully-connected graph, where red node
represents the global detection in this local assignment, while cyan nodes represents
local detections.}
\label{fig:tree_model}
\end{figure}

In our application we augment this tree model with additional edges from the major part (i.e., the neck) to
all other non-adjacent body parts.  This is illustrated in Fig~\ref{fig:tree_model}. 
Then, given the global detection associated with the neck, the conditional model is tree-structured
and can be optimized using dynamic programming in $O(|\mathcal{R}|k^2)$ time, where $k$ is the maximum number of detections per part ($k<15$ in practice). 

\section{Experiments}
\label{poseCellExp}

\begin{table}[t]
\begin{center}
\resizebox{.95\textwidth}{!}{
\renewcommand{\arraystretch}{1.5}
 \begin{tabular}{||c c c c c c c c c c | c||} 
 \hline
 Part & Head & Shoulder & Elbow & Wrist  & Hip & Knee& Ankle & mAP(UBody) & mAP & time (s/frame)\\ [0.5ex] 
 \hline\hline
 \textbf{Ours} & 93.3 & 89.6 & \textbf{79.8} & \textbf{70.1} &\textbf{78.8} &\textbf{73.2} &\textbf{66.6} &\textbf{83.2} &\textbf{79.1} &2.7 \\ 
 \hline
 \cite{NL-LMP} & \textbf{93.4} & \textbf{89.7} & 79.1 &68.6 &78.8 &72.5 &65.2 &82.7 &78.5 & \textbf{0.16} \\ \hline
 \cite{deepcut1} & 92.4 & 88.9 & 79.1 &67.9 &78.7 &72.4 &65.4 &82.1 &78.1
 &270* \\ \hline
\end{tabular}
}
\caption{We display average precision of our  approach versus the baselines for
  the various human parts as well as whole body. Running times are measured on
  an Intel i7-6700 quad-core CPU. Note that due to software and hardware
  limitations we cannot run \cite{deepcut1} on our own machine and thus we
directly cite the running time on validation set which was reported in their
paper.}
 \label{pose-estimation}
\end{center}
\end{table}

\begin{figure}[t]
\begin{center}
\begin{tabular}{ c | c c c}
\rotatebox[origin=c]{90}{Deeper Cut \cite{deepcut1}} &
\raisebox{-.5\height}{\includegraphics[width=0.275\textwidth]{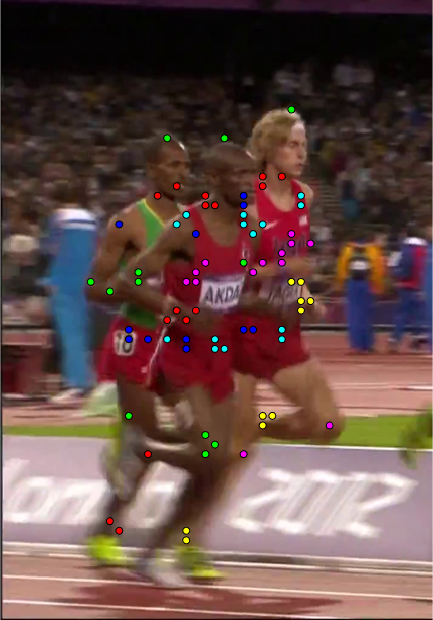}} &
\raisebox{-.5\height}{\includegraphics[width=0.275\textwidth]{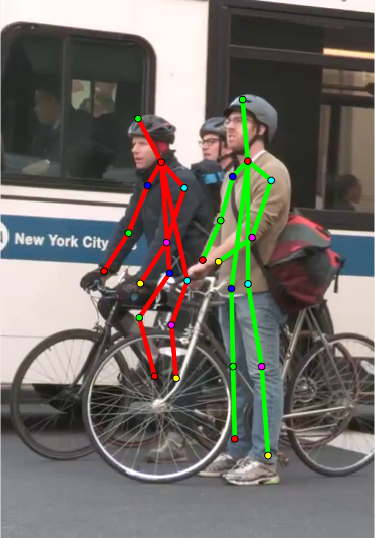}} &
\raisebox{-.5\height}{\includegraphics[width=0.27\textwidth]{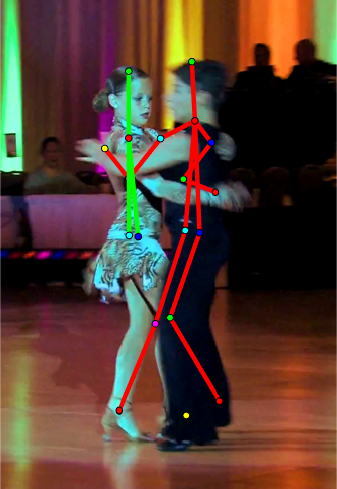}} \\
\rotatebox[origin=c]{90}{Our Approach} &
\raisebox{-.5\height}{\includegraphics[width=0.275\textwidth]{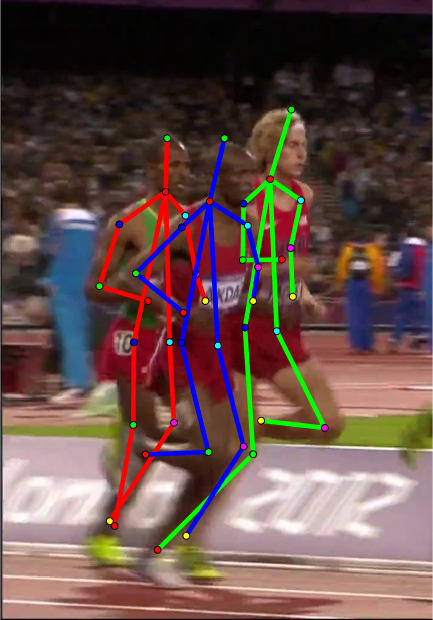}} &
\raisebox{-.5\height}{\includegraphics[width=0.275\textwidth]{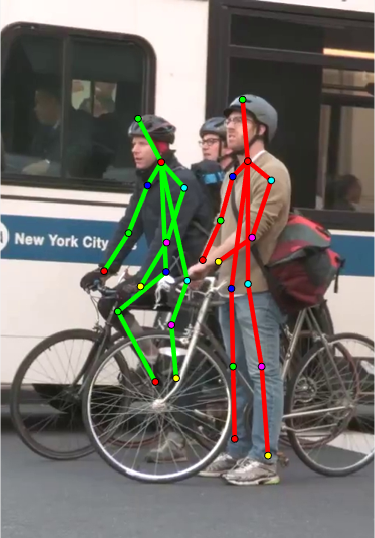}} &
\raisebox{-.5\height}{\includegraphics[width=0.27\textwidth]{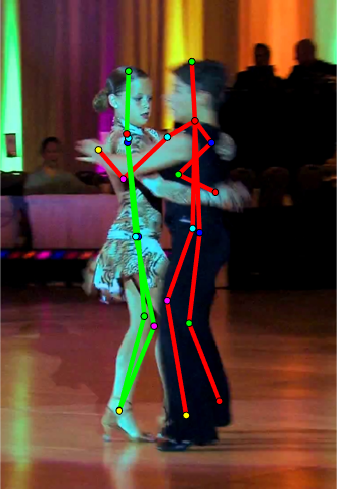}}
\end{tabular}
\caption{Qualitative comparison of \cite{deepcut1} (\emph{top row}) and our
  approach (\emph{bottom row}).  (\emph{Left column}) \cite{deepcut1}
  occasionally fails and produces many false positives per detection, while
  our approach avoid this by enforcing the fact that each individual person
  must have a neck. (\emph{Middle column}) We predict left knee of the
  person on the left better than \cite{deepcut1}. (\emph{Right column})
  \cite{deepcut1} fails to find the lower body parts of the person on the left
  and confuses ankle and kneel of the two people, while we successfully avoid
  these errors. }
\label{fig:quant}
\end{center}
\end{figure}

\subsection{Experiment Setup}
We evaluate our approach in terms of the Average Precision (AP) on the of
MPII--Multiperson training set~\cite{andriluka14cvpr}, which consists of 3844
images. For a fair comparison, we use the unary and pairwise costs directly
provided by Insafutdinov et al., and did not modify or weight these costs in any
way for any approach considered in this experiment.  Our model thus only differs
from \cite{deepcut1} and \cite{NL-LMP} in that our two-tier structure defines a
distinct and novel cost function.  In particular, our introduction of the
two-tier structure forces us to ignore the pairwise $\phi$ terms corresponding
to interactions between non-global detections that are associated with different
parts in a given pose.  A major benefit of this difference is a fast and
typically exact optimization process. Besides, local detections in a local
assignment often do not align well with the ground-truth position of a body-part
(\eg~Figure~\ref{fig:overview} and~\ref{fig:tree_model}), thus pairwise
interactions between such detections across part types can be noisy due to
inaccurate localization, and ignoring such interactions may contribute to more
accurate localization of body-parts.

In addition to the structure depicted in Figure~\ref{fig:tree_model}(a), we
found that adding additional edges for global pose that does not break
the conditional tree structure slightly improves Mean Average Precision (mAP)
from 78.8 to 79.1 with negligible increase in running time. The additional edges
we employ in our final model are left-hip to left-shoulder, right-hip to
right-shoulder and shoulders to head.

We set $\omega=30$ heuristically to discourage the selection of global poses
that include few detections, which tend to be lower magnitude in their cost.
After solving the LP \eqref{dualFormPose}, we tighten the relaxation if
necessary using odd set inequalities of size three
\cite{heismann2014generalization,Yarkony2017}, which does not interfere with pricing; more details can be
found in the supplements. 
In practice, however, we find that these refinements are rarely necessary to
produce integer solutions with identical cost to the LP relaxation at
termination.

We compare our results against two baselines: 1) \cite{deepcut1}, whose results
are obtained by its authors upon our request due to our limited acess to
computing resources and commercial LP solvers. 2) \cite{NL-LMP}, whose results
are obtained via running their code over the costs from~\cite{deepcut1}. We
found that employing the augmented-tree structure instead of a fully-connected
structure gives~\cite{NL-LMP} sligntly better performance (from 78.4 to 78.5).
Note that even based on the same graph structure,~\cite{NL-LMP} still has more
pairwise connections than our model as it considers connections between all
detections from different parts. 

\subsection{Benchmark Results}
As shown in Table~\ref{pose-estimation}, our approach runs much faster than
\cite{deepcut1} due to both the reduced model size and our more sophisticated
inference algorithm. While~\cite{NL-LMP} runs about 10x faster than our
approach, we achieve more accurate results than it: the improvement in mAP might
seem small (78.5 to 79.1), however we achieve much better AP on
difficult-to-localize parts such as wrist (70.1 versus 68.6) and ankle (66.6
versus 65.2), while we only use a subset of edges compared to~\cite{deepcut1}
and~\cite{NL-LMP}. Also keep in mind that all experiments are based on the same
set of unary/pairwise costs without any form of learning, thus our improvement
is solely due to our novel modeling for MPEE problem and the ability to find
global minimum of our cost.

We also note that the code of~\cite{NL-LMP} is in pure C++ and is heavily
optimized, while our code is in pure Python and we did not take advantage of the
parallelizable nature of our pricing problems. Nevertheless, we still achieve
considerable speed up over~\cite{deepcut1}. We will release the code and data we
used upon acceptance of this paper.

\section{Conclusion}
\label{concsectiondone}

We introduce a new formulation of the multi-person pose estimation problem, along with
a novel inference algorithm based on column generation that admits efficient
inference.  We compare our results to a state of the art algorithm and
demonstrate that our approach rapidly produces more accurate results than the
baseline. 

In future work we intend to apply our method to other domains where similar
local/global structure is present, and can assist in non-maximum suppression or
clustering, for example in relevant ILP optimization formulations of
multi-object tracking \cite{Tang2015}, moral lineage tracking\cite{moral}, and
MPPE tasks on video \cite{arttrack}.  

\bibliographystyle{ieee}
\bibliography{col_gen_bib}

\appendix
\section{Tighter Bound for Multi-Person Pose Estimation}
\label{tripPose}
A tighter LP relaxation than that in the main paper can be motivated by the
following observations: (1)  no more than one global pose can include more than
two members of a given set of three detections. (2) No more than one local
assignment can include more than two members of a given set of three detections
(either as local or global).  These constraints are called odd set inequalities
of order three \cite{heismann2014generalization}.  We formalize this below.

We refer to the set of all sets of three unique detections (triples) as
$\mathcal{C}$.  We use $C^{\mathcal{L}}\in \{ 0,1\}^{|\mathcal{C}| \times
|\mathcal{L}|}$ to define the adjacency matrix between triples and local
assignments.  Similarly we use $C^{\mathcal{G}}\in \{ 0,1\}^{|\mathcal{C}|
\times |\mathcal{G}|}$ to define the adjacency matrix between triples and global
poses.  Here $C^{\mathcal{L}}_{cl}=1$ if and only if  local assignment $l$
contains two or more members of set $c$.  Similarly we set
$C^{\mathcal{G}}_{cq}=1$ if and only if global pose $q$ contains two or more
members of set $c$.  We define $C^{\mathcal{L}},C^{\mathcal{G}}$ formally below.  
\begin{align}
&C^{\mathcal{G}}_{cq}=[(\sum_{d \in c}G_{dq})\geq 2] &\quad \forall c \in \mathcal{C}, q \in \mathcal{G}\\
\nonumber &C^{\mathcal{L}}_{cl}=[(\sum_{d \in c}L_{dl}+M_{dl})\geq 2] &\quad \forall c \in \mathcal{C}, l \in \mathcal{L}
\end{align}
\subsection{Dual Form}
We now write the corresponding primal LP for multi-person pose estimation with
triples added.
\begin{align}
\label{primalTriRelax}
 \min_{\substack{\gamma\geq 0 \\ \psi \geq 0 \\ G\gamma+L\psi \leq 1 \\ L\psi+M\psi\leq 1 \\-G\gamma+M\psi\leq 0 \\C^{\mathcal{G}}\gamma \leq 1 \\ C^{\mathcal{L}}\psi \leq 1}}\Gamma^\top\gamma+\Psi^\top \psi 
\end{align}
The constraints $C^{\mathcal{G}}\gamma \leq 1$  and $C^{\mathcal{L}}\psi \leq 1$
are referred to as ``rows" of the primal problem. We now take the dual of
Eq.~\eqref{primalTriRelax}.  This induces two additional sets of Lagrange
multipliers $\lambda^4,\lambda^5 \in \mathbb{R}_{0+}^{\mathcal{C}}$.  We now
write the dual below.
\begin{align}
\label{dualForwTrip}
\max_{\substack{\lambda \geq 0 \\ \Gamma+G^\top(\lambda^1-\lambda^3)+C^{\mathcal{G}\top}\lambda^4 \geq 0 \\
\Psi+L^\top\lambda^1+(M^\top+L^\top)\lambda^2\\
+M^\top\lambda^3 +C^{\mathcal{L}\top}\lambda^5 \geq 0}}  
-1^\top\lambda^1-1^\top\lambda^2-1^\top\lambda^4-1^\top\lambda^5
\end{align}
\subsection{Algorithm}
In order to tackle optimization we introduce subsets of
$\mathcal{C}^{\mathcal{G}}$ and $\mathcal{C}^{\mathcal{L}}$, denoted
$\hat{\mathcal{C}}^{\mathcal{G}}$ and $\hat{\mathcal{C}}^{\mathcal{L}}$
respectively.  These subsets are intially empty and grow only when needed.  We
write an optimization algorithm below in Alg \ref{cyccPOSE} with subroutines
$\mbox{ROW}(\gamma,\psi)$ (Section \ref{rowGenApp})  and
$\mbox{COLUMN}(\lambda)$ (Section \ref{colTripPosemake}) describing the
generation of new triples and columns respectively.  
 \begin{algorithm}[H]
 \caption{Column/Row Generation}
\begin{algorithmic} 
\STATE $\hat{\mathcal{G}} \leftarrow \{ \}$
\STATE $\hat{\mathcal{L}} \leftarrow \{ \}$
\STATE $ \hat{\mathcal{C}}^{\mathcal{G}} \leftarrow \{ \}$
\STATE $ \hat{\mathcal{C}}^{\mathcal{L}} \leftarrow \{ \}$
\REPEAT
\STATE $[\lambda]\leftarrow $ Maximize dual in Eq~\eqref{dualForwTrip} over column and rows sets $\hat{\mathcal{G}},\hat{\mathcal{L}},\hat{\mathcal{C}}^{\mathcal{L}},\hat{\mathcal{G}}^{\mathcal{L}}$
\STATE Recover $\gamma$ from $\lambda$
\STATE $\dot{\mathcal{G}},\dot{\mathcal{L}} \leftarrow \mbox{COLUMN}(\lambda) $
\STATE $ \dot{\mathcal{C}}^{\mathcal{L}},\dot{\mathcal{C}}^{\mathcal{G}} \leftarrow \mbox{ROW}(\gamma,\psi)$
%
\STATE  $\hat{\mathcal{G}}\leftarrow [\hat{\mathcal{G}},\dot{\mathcal{G}}]$
\STATE  $\hat{\mathcal{L}}\leftarrow [\hat{\mathcal{L}},\dot{\mathcal{L}}]$
\STATE   $\hat{\mathcal{C}}^{\mathcal{L}}\leftarrow [\hat{\mathcal{C}}^{\mathcal{L}},\dot{\mathcal{C}}^{\mathcal{L}}]$
\STATE   $\hat{\mathcal{C}}^{\mathcal{G}}\leftarrow [\hat{\mathcal{C}}^{\mathcal{G}},\dot{\mathcal{C}}^{\mathcal{G}}]$
 \UNTIL{ $\dot{\mathcal{G}}=[]$ and $\dot{\mathcal{L}}=[]$  and $\dot{\mathcal{C}}=[]$ }
\end{algorithmic}
\label{cyccPOSE}
\end{algorithm}
%
%
\subsection{Generating rows}
\label{rowGenApp}
Generating rows corresponding to local assignments is done separately for each
part. We write the corresponding optimization for identifying the most violated
constraint corresponding to a local assignment over a given part $r$ as follows.  
\begin{align}
\max_{\substack{c \in \mathcal{C}\\ R_d=r \; \; \forall d\in c }}\sum_{q \in \mathcal{L}} C^{\mathcal{L}}_{cl} \psi_q
\end{align}
Finding violated rows corresponding to global poses is assisted by the knowledge
that one need only consider triples over three unique part types as no global
pose includes two or more detections of a given part.  Hence only such triples
need be considered for global pose. For any given $c$ let the detections
associated with it be $c=\{ d_{c_1} d_{c_2} d_{c_3} \}$, the corresponding
optimization can then be written as below:
\begin{align}
\max_{\substack{c \in \mathcal{C}\\ R_{d_{c_1}}\neq R_{d_{c_2}}\\ R_{d_{c_1}}\neq R_{d_{c_3}}\\  R_{d_{c_2}}\neq R_{d_{c_3}}}}\sum_{q \in \mathcal{G}} C^{\mathcal{G}}_{cq} \gamma_q
\end{align}
Triples are only added to
$\hat{\mathcal{C}}^{\mathcal{L}},\hat{\mathcal{C}}^{\mathcal{G}}$ if the
corresponding constraint is violated.  
\subsection{Generating Columns}
\label{colTripPosemake}
Generating columns is considered separately for global poses and local
assignments.  The corresponding equations are unmodified from the main document
except for the introduction of terms over triples.
%
We write the IP for generating the most violated constraint corresponding to a
local assignment  given the global detection below. 

\begin{align}
\label{makeColTripL}
\min_{\substack{x \in \{ 0,1\}^{|\mathcal{D}|} \\x_{d_*}=1}}(-\lambda^1_{d_*}+\lambda^3_{d_*}-\theta_{d_*}) +\sum_{d \in \mathcal{D}}(\theta_d+\lambda^1_{d}+\lambda^2_d)x_d\\
\nonumber +\sum_{d_1,d_2 \in \mathcal{D}}\phi_{d_1d_2}x_{d_1}x_{d_2}+\sum_{c \in \mathcal{C}^\mathcal{L}}\lambda^5_{cl}[ \sum_{d \in c}x_{d}  \geq 2]
\end{align}

We optimize Eq.~\eqref{makeColTripL} via explicit enumeration as described in the
main paper.

For each $d_*$ such that $R_{d_*} \in \mathcal{R}'$ we compute the most violated
constraint corresponding to a global pose including $d_*$.  We write this as an
IP below.

\begin{align}
\label{makeColTripG}
\min_{ \substack{x \in \{ 0,1\}^{|\mathcal{D}|} \\ x_{d_*}=1\\  \sum_{d \in \mathcal{D}}R_{dr} x_d \leq 1 \; \; \forall r \in \mathcal{R}}}\sum_{ d\in \mathcal{D}} (\theta_{d}+\lambda^1_d-\lambda^3_d)x_{d} \\ 
\nonumber  +\sum_{d_1d_2 \in \mathcal{D}} \phi_{d_1d_2}x_{d_1}x_{d_2}+\sum_{c \in \mathcal{C}^\mathcal{G}}\lambda^4_{cq}[ \sum_{d \in c}x_{d}  \geq 2]
\end{align}

The introduction of triples breaks the structure of the problem, thus we can no
longer optimize Eq.~\eqref{makeColTripG} via dynamic programming.  We found that
employing the branch and bound algorithm proposed by~\cite{Yarkony2017} is not
computationally problematic for our problems as the number of triplets needed for
convergence is small.

\section{Additional Statistics for Results on MPII Training Set}
With up to 150 detections per image, we found our column generation solver
usually terminates with a few hundreds, and no more than 1000 columns (\ie~total
number of global poses and local assignments).

Out of all 3844 instances, we observe fractional LP solutions on 131 instances,
45 of which we successfully reached integer solutions with the help of triplets
constraints; for the rest of 86 fractional instances, it costs negligible
additional time to run trial version of CPLEX ILP solver to obtain integer
solutions given columns we generated.

\end{document}